# Matrix Profile XXVII: A Novel Distance Measure for Comparing Long Time Series

Audrey Der[1], Chin-Chia Michael Yeh[2], Renjie Wu[1], Junpeng Wang[2], Yan Zheng[2], Zhongfang Zhuang[2], Liang Wang[2], Wei Zhang[2], Eamonn Keogh[1]

{ader003, rwu034}@ucr.edu, {miyeh, junpenwa, yazheng, zzhuang, liawang, wzhan}@visa.com, eamonn@cs.ucr.edu

*Abstract*— The most useful data mining primitives are distance measures. With an effective distance measure, it is possible to perform classification, clustering, anomaly detection, segmentation, etc. For single-event time series Euclidean Distance and Dynamic Time Warping distance are known to be extremely effective. However, for time series containing cyclical behaviors, the semantic meaningfulness of such comparisons is less clear. For example, on two separate days the telemetry from an athlete's workout routine might be very similar. However, on the second day she might have changed the order in which she did push-ups and squats, added a few repetitions of pull-ups, or completely omitted dumbbell curls. Any one of these minor changes would defeat existing time series distance measures. Some "bag-of-features" methods have been proposed to address this problem; however, we argue that in many cases, similarity is intimately tied to the *shapes* of subsequences within these longer time series. In such cases, summative features will lack discrimination ability. In this work we introduce PRCIS, which stands for **P**attern **R**epresentation **C**omparison **i**n **S**eries. PRCIS is a distance measure for long time series, which exploits recent progress in our ability to summarize time series with "dictionaries". We will demonstrate the utility of our ideas on diverse tasks and datasets.

## I. Introduction

Distance measures are perhaps the most fundamental data mining primitives. Armed only with an appropriate distance measure, we can perform clustering, classification, anomaly detection, summarization, repeated pattern discovery, etc. Given the ubiquity and variety of time series, there are dozens of distance measures defined for this data type. However, almost all such measures are designed for short time series, say individual gestures, individual heartbeats, etc. Suppose instead we wish to address long time series. Surprisingly, there is very little work that considers such data. One solution advocated in [10] is to convert the long time series into a fixed feature vector, for example `A = [Mean(T), Skewness(T), Hurst(T)]`. Such an approach may be appropriate in some domains, however, in many cases the *shapes* of local subsequences may be the most discriminating features, and these are typically not well captured by global features.

Note that we cannot simply use classic time series distance measures. For example, DTW is rightly lauded for its invariance to out-of-phase local segments [27] . However, suppose we wish to compare two one-minute ECG segments from the same person. Because the heart rate is constantly drifting, it is likely that the two one-minute ECG segments will have a different number of beats, say 67 and 72. DTW is simply unable to "explain" the five extra beats, and will be forced to return a large distance suggesting the two traces are very dissimilar.

In addition, DTW (a special case of Euclidean distance) would not be suitable for long time series even if there was some mechanism that keep the periodicity fixed. Consider the example shown in Fig. 1.*left*. While each colored pair has the same periodicity, minor corruptions of the global trends are enough to thwart DTW's attempt to group similar pairs.

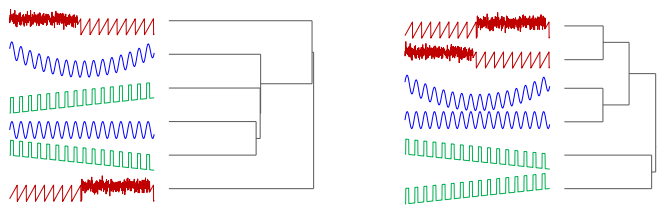

Fig. 1: Six time series, comprising of three obvious pairs, clustered using single linkage clustering. *left*) The clustering produced by DTW. *right*) the clustering produced by PRCIS, the method proposed in this paper.

The rest of this paper is organized as follows: in Section II we present our motivating observations. Section III describes the formal definitions and background and allows us to outline our approach. Section IV contains an extensive experimental evaluation. Finally, we offer conclusions and thoughts on future directions in Section V.

## II. Motivation and Related Work

Because we have used the term "long time series" without defining it, we repair that omission now. For our purposes, "long" does not necessarily correlate with the number of datapoints. For example, a single heartbeat may be a second long, and represented with 128 datapoints. However, some ECG apparatus can record at up to 32,768 Hz. In a sense, a single heartbeat with either 128 or 32,768 datapoints covers the same wall clock time and has the same intrinsic dimensionality. By long time series we mean a time series that has at least hundreds of features, such as peaks and valleys.

While there are excellent distance measures for time series, such as Euclidean distance and DTW, these measures are suited to atomic subsequences, for example, comparing two heartbeats, or two gestures, or two days of road traffic density. Note that the UCR/UEA Archive [6] *exclusively* contains datasets of this type. The top performing methods on that archive include deep learning and non-deep learning based methods, including: HIVE-COTE [15], ROCKET [7], TS-CHIEF [23], InceptionTime [8], BOSS [22] (a dictionary learning algorithm), and a good baseline in ResNet [24].

---

In contrast, we desire a distance measure that can compare items at a higher semantic level, for example comparing two hour-long ECG traces, before and after a medical intervention, or comparing two 90-minute soccer games, or comparing road traffic density of two freeways over a year, etc.

It should be clear that neither Euclidean distance nor DTW will work well if directly applied in such circumstances. For example, two one-hour traces of walking gait may have a differing number of steps. DTW can compare two steps that are locally out of phase, but it cannot map say ten steps to eleven steps, and that single unexplained step will completely swamp any similarity that may or may not exist.

Likewise, imagine comparing the road traffic density of two freeways over a year. We imagine two freeways could have near identical traffic patterns, but one of them comes from a suburb that has a slowly increasing population. This slight linear trend difference might be imperceptible over a day or a week, the typical scale considered by Euclidean distance, but over a year the small linear trend difference will again completely swamp any similarity that may or may not exist.

These issues are understood in the community, but to date we are unaware of any automatic solution. Attempts to mine such data typically involve a lot of human intensive preprocessing steps, cleaning the data (detrending, normalizing, imputation of missing values, etc.), and the manual extraction of clean and representative examples to feed into downstream algorithms.

Our basic plan is to create a compact dictionary for each time series and calculate the distances between each pair of dictionaries. The term *dictionary creation/learning* is somewhat overloaded in computer science. It is often a synonym for sparse coding, a class of representation learning methods which aims at finding a sparse representation of the input data in the form of a linear combination of basic elements as well as those basic elements themselves. The elements of the representation learned do not typically come from the data. In contrast, we propose to use dictionaries that comprise of elements from the data itself.

### III. DEFINITIONS AND BACKGROUND

We begin by defining the key terms used in this work. The data of interest here is *time series*.

DEFINITION 3.1. A time series $T$ is a sequence of $l$ real-valued numbers $t_i$: $T = [t_1, t_2, \ldots, t_l]$.

Most data mining algorithms do not operate on the entire time series, but instead consider only local *subsequences* of the time series.

DEFINITION 3.2. A *subsequence* $T_{i,j}$ of a time series $T$ is a continuous subset of data points from $T$ of length $j - i$ starting at position $i$. $T_{i,j} = [t_i, t_{i+1}, \ldots, t_{i+j-1}], 1 \leq i \leq l - j + 1$.

The length of the subsequence is typically set by the user based on domain knowledge. For example, for most human actions, ½ second may be appropriate, but for considering traffic congestion patterns, subsequences of one or two days may be more fitting.

As noted above, the intuition for our solution is to create a compact dictionary for each time series and calculate the distances between each pair of unique dictionaries. Thus, we will next examine some of the ways in which the dictionary could be created.

#### A. Dictionary Creation

There are at least four ways a dictionary of short time series patterns could be constructed from a long time series.

- **Yeh's Dictionary** [26]: Consider a time series $T$ with conserved patterns, supply a window length $w$ and the number of patterns $n$ desired for the resulting dictionary. Computing the matrix profile MP of $T$ will reveal the locations of these motifs. Building the distance profile of the motif $Q$ against $T$ will indicate the locations of $Q$ and subtracting this distance profile from MP will remove all locations of $Q$ from MP. Repeating this process $n$ times will produce up to $n$ patterns. Unless otherwise stated, this is the dictionary creation algorithm we use in this work.
- **Random Dictionary**: This is building a dictionary by randomly extracting elements of a long time series. While meant to be a naïve baseline, random sampling is known to be competitive for some tasks.
- **Time Series Snippets**: A snippet is an unsupervised time series primitive that rewards both fidelity and coverage [13]. Given a time series $T$, the Snippet Selection Algorithm captures "typical" behavior in an ordered list of substrings. Other research efforts produce similar compact summarizations of time series, for example BeatLex (Beat Lexicons for Summarization) [11].
- **Calendar Dictionaries**: This is an option only for datasets constrained by human circadian patterns (e.g., traffic density, network usage, etc.). Someone familiar with the domain can hand curate a dictionary by selecting a handful of informative days.

While our proposed distance measure is agnostic to the dictionary creation method used, unless otherwise stated we will use Yeh's Dictionary in this work. Regardless of how the dictionary is created, we represent a dictionary of a single time series as a *dictionary exemplar*.

DEFINITION 3.3. A *dictionary exemplar* (henceforth simply referred to as a dictionary where there is no confusion) refers to any dictionary produced by a dictionary learning method where the input is a time series $T$ and has an output of a dictionary $T_d = \{p_1, p_2, \ldots, p_n\}$ where there are $n$ patterns.

We refer to the values for the notation in Fig. 2 to be the size of the dictionary $S$ with patterns of length $L$. $S$ and $L$ are simply the parameters passed to the dictionary creation algorithm. Suppose a random dictionary algorithm created a $T_{d[S,L]}$ such that it extracted $S$ patterns of $L$ length as random non-overlapping subsequences from $T$. In this case, the notation is true to value. In contrast, Yeh's Dictionaries extract and then merge patterns with overlapping subsequence indices, making $S$ an upper bound and $L$ a lower bound.

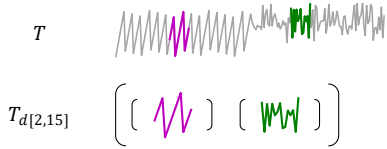

Fig. 2: *top*) A time series *T* with two different semantic patterns, a sawtooth and a noisy signal. *bottom*) A dictionary that summarizes *T*, with a cardinality of two with patterns of length 15, denoted as $T_{d[2,15]}$.

### B. Desirable Properties of a Distance Measure

Before defining a distance measure for dictionaries, it will be instructive to consider desirable properties for such a measure. To allow us to be concrete, we will use discrete strings as proxies for time series. The reader will appreciate that continuing this analogy, the Hamming distance is an excellent proxy for Euclidean distance. For example, the Hamming distance `d(cat,bat)=1`, represents two quite similar time series with a low Euclidean distance.

With the exception of calendar-based dictionaries, it is obvious that we cannot expect dictionaries to be phase aligned. For example, suppose we have two similar long series:

```
A = …catcatcatcatdogdogdogcatcat…
B = …dogdogdogcatcatcatdogdogdog…
```

If we use Yeh's algorithm to create two dictionaries from these datasets, we may obtain:

`A_d={cat,dog}`, `B_d={atc,dog}`

We do not wish to penalize the apparently great distance between "cat" and "atc", as we realize that is just the result of the vagaries of the dictionary building algorithm. We can be invariant to this by holding one word fixed, and comparing to every circular shift of the other word, equivalent to:

`RI_dist(cat,atc)=min([d(cat,atc),d(cat,tca),d(cat,cat)])`

Here, `RI_dist` is a phase invariant distance or "k-shape", computed in $O(mlog(m))$ [20].

There is another issue that the dictionary creation algorithm may throw at us. Revisiting examples *A* and *B* above, we may be presented with:

`A_d={cat,dog}`, `B_d={dog,cat}`

Dictionary building algorithms may not be consistent in ordering patterns with respect to frequency or representativeness, and two time series could be very similar, but produce dictionaries with words in different orders. For example, because Bangkok and Lima are near perfect antipodes, they have similar climates, but exactly six months out of phase. Once again, this is easy to handle. For *each* word in dictionary *A*, find its nearest neighbor in dictionary *B*, and vice versa.

```
d_dict(A_d,B_d) = {min(RI_dist(cat,dog),
  RI_dist(cat,cat)) + min(RI_dist(dog,dog),
  RI_dist(dog,cat))} + {min(RI_dist(dog,cat),
  RI_dist(dog,dog)) + min(RI_dist(cat,cat),
        RI_dist(cat,dog))}
```

The first pair of curly braces is the computed distance from `A_d` to `B_d`, and the second pair is the computed distance from `B_d` to `A_d`. It is important to note that `d_dict` is not the functionality of PRCIS, but the expression is useful to communicate the intuition behind PRCIS.

For flexibility, we desire the capability to compare patterns of varying lengths and dictionaries of varying cardinalities. Given two more time series:

```
C = …catcatcatcatdogdogdogcowcowcow…
D = …bobcatbobcatodogdogdobobcatbob…
```

This now gives us four time series dictionaries to consider:

`A_d={cat,dog}`,`B_d={dog,cat}`,

`C_d={cat,dog,cow}`,`D_d={bobcat,cat}`

The issue of varying cardinalities is trivial; we simply have more `RI_dist` computations inside the curly brace pairs. Likewise, handling varying pattern lengths is done by generalizing the rotation invariant distance to allow strings of different lengths. We can do this by taking the longer of the two strings and concatenating it to itself, then using the shorter string as a query, finding the best substring within the longer string.

`RI_dist(bobcatbobcat,cat) = d_substr(bobcatbobcat,cat)`

Using these ideas, the distance between `{cat,dog}` and `{bobcat,cat}` will be zero. This may initially be unintuitive until you review the strings that originally produced them, and recognize that they could be very similar:

```
A = …catcatcatcatdogdogdogcatcat…
D = …bobcatbobcogdogdogdbobcatbo…
```

We are almost done, but there is one last desirable invariance. It may happen that while two dictionaries are generally very similar, one of them has one or more words that are unique. For example, imagine we have `X_d={cat,dog,cow}`, `Y_d={cow,cat,dog,%$@^*}`. The unusual word in `Y_d` may have many causes. Perhaps the corresponding sensor had a long disconnection artifact that the dictionary algorithm thought worthy of representation. Although these two dictionaries are near identical, the single strange pattern will dominate the overall distance, and mask the similarity. Our solution is to forgo the brittle mean distance (between each word and its rotational invariant nearest neighbor in the other dictionary), but instead use the median distance. The median distance is much more forgiving of a handful of words that have no obvious match.

This section has been somewhat long and detailed, however with these ideas in mind, the exposition of the real-valued PRCIS distance measure in the next section is straightforward.

### C. PRCIS Dictionary Distance Measure

PRCIS stands for <u>P</u>attern <u>R</u>epresentation <u>C</u>omparison <u>i</u>n <u>S</u>eries and is the distance measure by which we propose to measure the similarity between two arbitrary real-valued dictionaries. We outline PRCIS in TABLE 1 and TABLE 2.

TABLE 1: THE ALGORITHM FOR PRCIS.

```
Algorithm: PRCIS
Input: dictionary A, dictionary B
Output: dictionary-distance-measure
1  AtoB ← PRCIS_AtoB(A,B)
2  BtoA ← PRCIS_AtoB(B,A)
3  all_dists ← concat(AtoB,BtoA)
4  return (median(all_dists))²
```

We discuss TABLE 2 first, as it contains the bulk of the logic in PRCIS. TABLE 2 finds the nearest neighbors for every pattern in dictionary *A* from the patterns in dictionary *B*. For every element in *A*, its distance to every element in *B* is calculated in part by running MASS [18]. Between the pairs of patterns, the shorter pattern is treated as the query, *q*, and the longer pattern is concatenated to itself and treated as the time series, *ts* (lines 5-7). In lines 8-12, the minimum of the minimums of these distance profiles is recorded as the distance from that pattern in *A* to its nearest neighbor in *B*. This occurs twice.

The second time PRCIS_AtoB is called: for every pattern in *B*, calculate the distance to every element in *A*. When this is done, lines 3-4 of TABLE 1 of PRCIS return the square of the median of this list of distances, and this represents the distance between the dictionaries *A* and *B*.

TABLE 2: A SUBROUTINE CALLED IN PRCIS (TABLE 1).

```
Algorithm: PRCIS_AtoB
Input: dictionary A, dictionary B
Output: nn-distances-from-A-to-B
1    dists ← []
2    for pA ∈ A do
3        nn_dist ← ∞
4        for pB ∈ B do
5            l,q ← pA,pB # switch if len(pB)>len(pA)
6            ts ← concat(l,l)
7            min_dp ← min(abs(MASS(ts,q)))
8            if min_dp < nn_dist do
9                nn_dist ← min_dp
10           end if
11       end for
12       dists.append(nn_dist)
13   end for
14   return dists
```

The resulting distance is a measure, but not a metric. However, [4] and others have forcefully argued that in some cases the metric property of triangular inequality must be violated to produce sensible results. Moreover, the ubiquitous DTW is also not a metric, but regarded as SOTA for comparing short time series [10][27].

### D. PRCIS Distance Measure Complexity

We say $n$ is the largest number of words in any dictionary, and the number of dictionaries being compared is $m$. Generally, $m$ is nontrivially large and $n \ll m$. MASS is proven to be extremely fast [18] (progressively so with newer versions), so we abstract its runtime to $O(1)$. The time complexity of PRCIS is $O(m^2 n^2)$. Additionally, the runtime of MASS only depends on the length of patterns. The number of dictionaries to compare or the cardinality of each dictionary have a greater effect on runtime. Memory is not a bottleneck for this distance measure.

## IV. EMPIRICAL EVALUATION

To ensure our experiments are reproducible, we have created a supporting website [21] containing all data, code, and results. Additionally, it houses other experiments and comparisons omitted here for brevity.

### A. Clustering

We begin by showing the utility of our algorithm for clustering, as this allows the most direct and visually intuitive demonstration of the invariances that our measure achieves. The full set of combinations of hierarchal clustering parameters (average, single, or complete linkage under L1, L2, or cosine similarity) produce similar clusterings are available at [21]).

### 1) Electrical Power Demand

We consider a three-year long dataset of electrical power demand from European countries [19]. As Figure 3 shows, the data are surprisingly diverse.

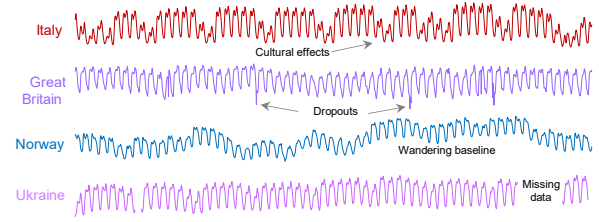

Fig. 3: Two-month subsequences of the electrical power demand data from four countries in Europe.

There are at least three sources of variability. The first is cultural; many Mediterranean countries regard weekends as sacrosanct, and they have a visible decrease in power required for weekends, as many businesses close (see Italy in Fig. 3).

There are also obvious weather effects, the Mediterranean climate has low variability, but countries in northern Europe have a power demand that clearly reflects the vagaries of the climate (see Norway in Fig. 3). Finally, there may also be policy effects. For example, some countries incentivize off peak consumption, however this is implemented differently in different countries. For example, Denmark, Norway, and Sweden use spot-market-based pricing, whereas Estonia, Spain and the UK use dynamic real-time pricing [14].

Fig. 3 further hints at why this dataset is challenging from most clustering algorithms. There are regions of missing data and dropout and spikes that seem to be meaningless (at least to the task at hand) but are likely to confuse most algorithms. Nevertheless, Fig. 4 shows a successful clustering.

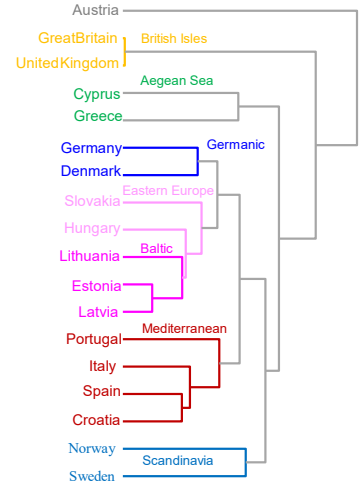

Fig. 4: The electrical power demand data from 25 of the countries under complete linkage hierarchical clustering using $T_{d[4,48]}$, where 48 datapoints represent two days. We opted *L*=48 to cover the transition from weekends to weekdays (and vice versa) and *S*=4 to cover types of days we experience, such as a *weekday* (Monday–Thursday), *weekend, holiday,* and *Fridays*.

This dataset spans from December 31, 2014 23:00:00 GMT to September 30, 2020 23:00:00 GMT. The results in Figure 4 are very intuitive, reflecting the major regions of Europe. A handful of placements are debatable. Denmark is a Scandinavian country; however, it does share a 68-kilometer border with Germany but is only connected to Sweden by a bridge.

While the overall clustering is very intuitive, Austria is a surprising outlier that we might otherwise have expected to group with Germany. We can only speculate as to the reason. Austria is famous for having "golden Sunday", a tradition (formally enshrined in federal law in 2003 [25]) for all shops to be closed on Sunday. This deference to the sabbath has all but disappeared in the rest of Europe. In any case, this is exactly *why* we do such clusterings, to discover interesting and unexpected findings that can be further investigated.

We attempted to achieve similarly intuitive results with other methods, including k-shape [20], Catch22 [16], and obvious strawmen, such as clustering random days that are not an obvious anomaly. None of these attempts yielded a clustering that was significantly better than random. For brevity, we report these experiments in [21].

*2) Business Merchants*

Identifying a merchant's true business type is vital to ensure the integrity of payment processing systems, as a merchant may intentionally or unintentionally report a false business type to the payment process network [28]. When representing merchants with hourly aggregated time series, time series analysis is effective for identifying a merchant's business type [28]. For example, toll collections peak during rush hours and meal delivery services are busy around lunch and dinner hours.

Thirty-eight merchants from five different categories (vending machines, online videogame retailers, toll collection, parking meters, and meal delivery services) are sampled for this study. The names of the merchants are not released to maintain confidentiality. For each merchant, we generate the number of transactions per hour from January 1, 2018 to June 30, 2021. In Fig. 5, we show an example of two vending machines.

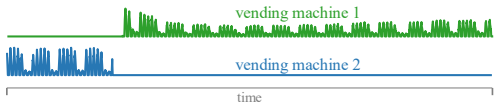

Fig. 5: Vending machine 1 may be newly added, and vending machine 2 may be recently removed. Computing the distance between them from a *global* perspective is doomed to fail.

In Fig. 6 we compare Yeh's Dictionaries + PRCIS with three baselines, visualizing each merchant as a coordinate computed using t-SNE. Our method and baselines' problem constructions are as follows:

- **Yeh's Dictionaries + PRCIS:** We construct a $T_{d[18,168]}$ for each series (patterns of $L$=168 span one week). If Yeh's dictionary algorithm returns a $T_d$ with the maximum 18 patterns allowed, the total length of the patterns in $T_d$ will be no longer than 10% of the length of the full series.
- **Entire Series + DTW:** The series are z-normalized and the DTW distance is computed pairwise between subsequences of the full series, each 182 weeks long.
- **Random 10% + DTW:** A random subsequence 10% of each time series length are extracted, then z-normalized and the DTW distance is computed pairwise between subsequences. This baseline is to determine if PRCIS' success is due to using only a subset of the original series.
- **Systematic Random 10% + DTW**: Similar to Random 10%, each merchant is also represented with a subsequence in lieu of the full series. However, all merchants use subsequences covering the same period of time.

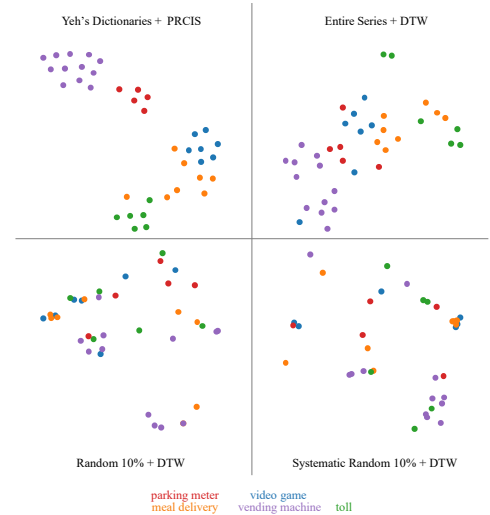

Fig. 6: Four different combinations of inputs and distance measures. The scatterplot associated with PRCIS shows a clear separation of merchants from the five business types.

PRCIS successfully captures the business type of the merchants, attending to local subsequences at a meaningful scale (one week). The other baselines look at each series from a much longer time horizon. Series like those in Fig. 5 have long "no signal" subsequences, and a distance measure will fail if unable to ignore these.

*3) Milling Machine Table Acoustic Emissions*

The previous datasets were relatively low frequency datasets. Here we consider data sampled at 250Hz. The NASA/Berkeley Milling Dataset [2] contains acoustic emission from a milling machine under different depths of cuts, feeds, and materials used. In Fig. 7.*top* we see that these cases are difficult to visually distinguish.

It might not be obvious why we need a dictionary-based method here. However, the documentation that accompanies the data makes it clear that the data is polymorphic. There is data corresponding to three phases: *entry* cut, *steady* cut, *and* exit cuts (although the boundaries between these behaviors are not explicitly given). In addition, there may be differences due to the direction of motion relative to the cutting head rotation, i.e., climb cut vs. conventional cut, although this is not documented.

TABLE 3: THE CLASSES AND CONDITIONS OF THE SERIES IN Fig. 7.

| Case | Depth of Cut | Feed | Material |
|---|---|---|---|
| 1 | 1.5 | 0.5 | Cast Iron |
| 3 | 0.75 | 0.25 | Cast Iron |
| 7 | 0.75 | 0.25 | Steel |
| 16 | 1.5 | 0.5 | Steel |

Using PRCIS, we can achieve the clustering in Fig. 7.*bottom*, where each of the series is 9,000 datapoints,

equivalent to 36 seconds of wall clock time. In Fig. 8 we cluster the same series using the Catch22 framework [10].

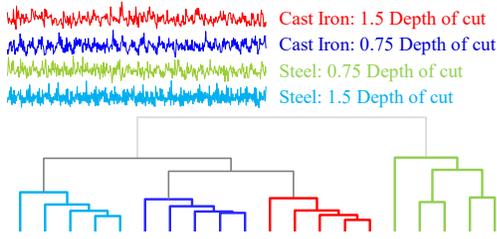

Fig. 7: *top*) Two-second-long samples from four semantic classes in a machining dataset. *bottom*) The average linkage clustering of twenty such time series under $T_{d[4,512]}$.

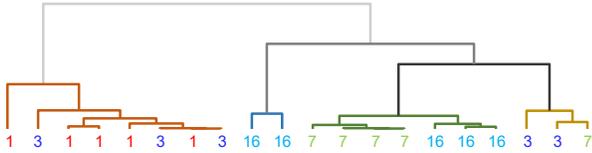

Fig. 8: Catch22 under complete linkage clustering on the same time series as those in Fig. 7.

Because this dataset is of both uniform length and has an available ground truth, we can make a third comparison in k-Shape [20], which produced the following four clusters:

```
['3' '3' '3' '3' '3' '7']
['1' '1' '1' '1' '1' '7']
['16' '16' '16']
['7' '7' '7' '16' '16']
```

These clusters are more homogenous than those produced by Catch22, but do not distinguish cases 7 and 16 nearly as well. Additional comparisons of clusterings based on spectral features are reported at [21].

### B. Classification

Here we turn our attention from clustering to classification. There are literally thousands of papers on classification of short time series, say individual heartbeats or gestures. (In essence, every paper that cites the UCR classification archive). However, these works assume that individual heartbeats/gestures/events have been extracted from the data streams. In many cases, this *extraction* may be a more difficult task than the classification problem itself. As the clustering examples we have considered hint at, we are interested in much longer, unstructured datasets in domain agnostic settings. There is very little literature on time series classification under these assumptions.

Here we consider two very different long time series classification challenges. While this is a subjective claim, we believe that in general neither of these tasks could be solved by human inspection. For example, for USC-HAD [29] `Walking-Forward` and `Walking-Upstairs` are visually very similar. To avoid over tuning the algorithm, we spent a few minutes "playing" with a tiny subset of dataset to select *S* and *L*. For each algorithm considered, we used 1NN Leave-One-Out (LOO) classification to evaluate its performance.

Despite that the classification works we reference in Section II are trained on single-event time series, we compare to ROCKET for completeness. ROCKET is the fastest of the SOTA mentioned in Section II and executes in approximately the same time scale as Catch22 and PRCIS ("minutes to hours" as opposed to "hours to days"). The out-of-the-box implementation from the authors [7] does require a uniform length time series dataset. We offer results using zero-padding and padding any shorter series to itself, clipping to the length of the longest series. As aforementioned, WeAllWalk series lengths varied from lengths 1,525-7,725 datapoints, and USC-HAD series lengths varied from 600-13,500 datapoints.

The feature-based approach we compare to is Catch22. Catch22's best subset of features was selected through simple greedy forward selection. If an input time series results in a Catch22 feature vector containing any NaN values, the series is simply discarded, and not counted against Catch22. This situation was not applicable to WeAllWalk, but in USC-HAD, 28 such series were discarded.

In the case of Random Dictionaries + PRCIS, reported results are the average of 10 runs. Six nonoverlapping subsequences of length 150 were selected randomly. In the cases where time series may be shorter than *S*L* (in this case, *S*L* = 900), we opt to include as many patterns of length *L* as possible. For example, no more than four patterns can be extracted from a series of length 600 when *L*=150. By sight, the smallest unit of cycles seemed to appear in lengths of 50 datapoints, so we opted to select a pattern length following a factor of 50.

We compared two variants of PRCIS. The random dictionary version offers evidence for our claim that using a *subset* of the data can be better than using *all* the data, and the Yeh's Dictionary variant offers evidence that well-chosen subset of the data is better again. TABLE 4 summarizes the results.

TABLE 4: THE LOO ACCURACY OF SEVEN APPROACHES ON THREE DATASETS.

| *Dataset:* | USC-HAD | WeAllWalk |
|---|---|---|
| *Dataset Size:* | 630 | 42 |
| *Default Rate:* | 11.11 | 28.75 |
| *Dictionary Parameters:* | $T_{d[6,150]}$ | $T_{d[8,25]}$ |
| ROCKET, zero-padding | 0.0 | 28.57 |
| ROCKET, self-padding | 0.0 | 28.57 |
| Catch22/ Best one feature | 36.67 | 76.19 |
| Catch22/ All features | 44.13 | 66.67 |
| Catch22/ Best feature subset | 64.92 | 97.62 |
| PRCIS, Rand. Dict. | 71.08 | 92.14 |
| PRCIS, Yeh's Dict. | **73.33** | **100.00** |

Not listed in TABLE 4, we reference the reported results on USC-HAD by [1] using handcrafted features and a tuned multilayer perceptron to obtain an accuracy of 74.71%. In contrast, we have opted to not preprocess the dataset to remove spurious subsequences or extract individual gait cycles. Additionally, many papers edit the classes considered in nonstandard ways. We have set nothing except the pattern length to 150 based on a quick visual inspection of a few traces.

### C. Effect of Dictionary Size

Classification experiments are the most direct way to evaluate the effect of dictionary size on PRCIS's ability to effectively represent long time series. While there is a single semantic label for each class, i.e., "`Walking-Forward`", our claim is that for a long time series, it is unlikely that any single subsequence can represent the concept. For example, even the apparently monolithic "`Walking-Forward`" may actually consist of multiple regions, *getting-up-to-speed*, *cruising-at-constant-speed*, etc. If our assumption is true, we would expect to see poor results for the smallest dictionary size and increasing (but with diminishing returns) accuracy as we make the dictionaries larger. In Fig. 9 we repeat the experiment on the USC-HAD dataset shown in TABLE 4.

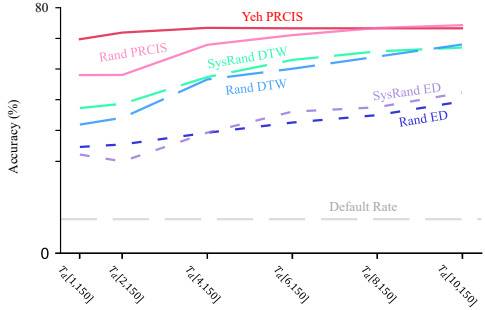

Fig. 9: The dictionary size vs. accuracy on the USC-HAD over different pairs of dictionary algorithms and distance measures.

Largely, the results in Fig. 9 support our hypothesis: while a dictionary of size one is significantly better than default rate random guessing, additional words do improve accuracy. However, the improvements yield diminishing returns as the larger dictionaries begin to fully represent the diversity of the behavior. In the case of Yeh+PRCIS, passing in a greater value for *S* seems to have little effect after six patterns. This is because most of the behaviors are relatively homogamous. Datasets with more diverse behaviors (i.e., the electrical power demand data of Section IV.A.1 with its different cultural and weather seasons) may require larger dictionaries.

### D. Towards PRCIS Anomaly Detection

There has been a recent explosion of interest in time series anomaly detection. A wealth of techniques have been applied to this task (including deep learning methods [3]). However, there is increasing evidence that simple distance-based methods are competitive in this area. For example, a recent paper compared over a dozen variants of SOTA deep learning approaches to a simple distance-based method (discords as represented by the Matrix Profile [13]) on 250 datasets, and discovered that the distance-based method was significantly better [3].

However, this result only addresses short anomalies, a single arrythmia or a single stumble in a long normal walk. Our warnings about the appropriateness of techniques for short time series being blindly applied to long time series apply here, perhaps even more so than for clustering or classification. In particular, we claim that there are some problems for which essentially all techniques in the literature are inappropriate. To demonstrate this, consider the following example. The example is slightly contrived but uses real data [5], and reflects a real industrial problem.

The performance of electrical motors is often monitored with accelerometers. Some types of motor anomalies are easy to detect from such data, for example a faulty bearing or worn brushes. However, there are much subtler patterns that may be anomalous. For example, an S3-class industrial motor is designed for intermittent periodic duty, a sequence of cycles containing periods of constant loads and (typically) a period at rest. There may be periods of different levels of constant loads; a machine may cycle between *rest*, *high* load (to raise a conveyor), *low* load (to lower the conveyor), back to *rest*, in an cycle. The normal behavior of this device is polymorphic, no single time series shape can represent all the version cycles.

It may be possible to build an anomaly detector in this domain with detailed domain knowledge. However, we propose to address this problem with a simpler approach. We propose to build a dictionary from a sample of normal data. We will not "tell" the algorithm where the different cycles begin (in any case, we generally may not know this), instead we assume that the dictionary creation algorithm will automatically choose representative patterns, so long as the dictionary size is greater than or equal to the true number of atomic behaviors. In Fig. 10 we have done exactly this on an example of normal behavior of an industrial paper shear for a large volume book printer.

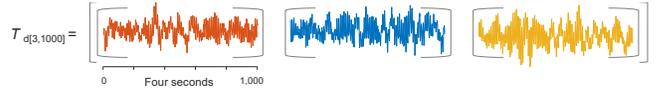

Fig. 10: A dictionary $T_{d[3,1000]}$ learned from a trace of normal behavior from an industrial shear.

In this case, the data is so complex that it is impossible to visually confirm that it represents the diversity of duty cycles in the industrial process. However, we can test this. In we create a PRCIS *distance profile*, "sliding" each pattern across an eleven-minute trace from the same motor. Because PRCIS (TABLE 1) is defined as a distance measure between a *dictionary* and a *dictionary*, we briefly cover the PRCIS distance profile in TABLE 5, which is defined as the distance from a *dictionary* and a *time series*.

The data in Fig. 11.*top* has a 32-second anomaly starting five minutes in. Gratifyingly, the PRCIS distance profile Fig. 11.*bottom* peaks at the location of the anomaly, caused by one of the cycles drawing a higher load due to a paper jam.

TABLE 5: THE ALGORITHM TO COMPUTE THE PRCIS DISTANCE PROFILE.

```
Algorithm: PRCISDistProf
Input: dictionary D, time series T
Output: dictionary-to-series-distance-measure
1  dps = []
2  for p_d in D:
3    dps.append(MASS(T,p_d))
4  meta_dp = elementwise_mean(dps)
5  PRCISdistprof = movemean(meta_dp,L)
6  return PRCISdistprof
```

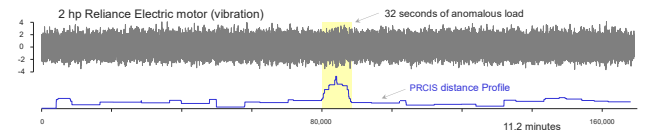

Fig. 11: *top*) An eleven-minute trace of motor powering an industrial paper shear. *bottom*) The distance between the dictionary shown in Fig. 12 to local subsequences of length 1,000 measures with the PRCIS distance measure. This curve was smoothed for clarity, not affecting the result. The parameters for smoothing were a factor of *L*.

Before continuing discussion, we want to ward off a possible misunderstanding. PRCIS is designed for long time series. Here long does not refer to the data being monitored, in any case that is data normally assumed to be effectively unbounded. Instead, long here refers to the length of the subsequences being considered. For example, the words in Fig. 10 are much longer than the length of subsequences considered in all anomaly detection studies we are aware of [3][13].

We have shown that PRCIS can detect (or at least peak at) at the anomaly; can other approaches? In Fig. 12 we consider the Matrix Profile [13], and Telemanom [12], a widely cited deep learning approach. The Matrix Profile was computed using a subsequence window of 256. The Telemanom Profile is an error curve under exponential weighted moving average smoothing (EWMA smoothing).

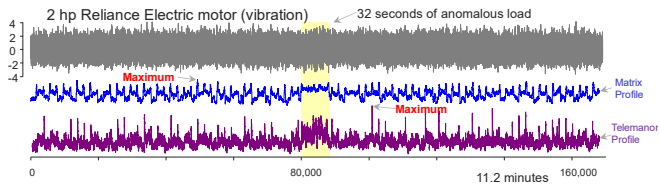

Fig. 12: *top*) An eleven-minute trace of motor powering an industrial paper shear. *bottom*) The distance profiles for two anomaly detection algorithms peak in the wrong places.

In fairness to the inventors of Matrix Profile and Telemanom, while they do not explicitly state this, it is clear that they intended their algorithms to work with short time series subsequences, not the longer semantically diverse subsequences and regions we are consider here.

This example is somewhat tentative and speculative; unlike the more common "short anomaly" problem, there are currently no standard benchmarks to evaluate this task. However, we believe that this example shows the promise of considering anomaly detection problems at the higher level of combinations of behavior, rather than a single irregular shape.

V. CONCLUSIONS AND FUTURE WORK

We have introduced PRCIS, a novel distance measure for comparing long time series. We have shown that using PRCIS can produce intuitive and semantically correct clusterings and embeddings of data. Moreover, it can do this in the presence of noise, spikes, dropouts, and even missing data. In future work we plan to consider further downstream uses of PRCIS, and investigate ways to set its two parameters automatically, probably by exploiting recent advances in using Minimum Description Length (MDL) for time series [11]. We have made our code and datasets freely available in perpetuity, to allow others to confirm and extend our findings.

ACKNOWLEDGMENTS

Our thanks to the donors of datasets. We also thank Dr. Greg Mason.

REFERENCES

[1] Reem Abdel-Salam, et al. (2021) Human activity recognition using wearable sensors: review, challenges, evaluation benchmark
[2] Alice Agogino and Kai Goebel (2007). BEST lab, UC Berkeley. Milling Data Set, NASA Ames Prognostics Data Repository (ti.arc.nasa.gov/project/prognostic-data-repository).
[3] Julien Audibert, et al. From Univariate to Multivariate Time Series Anomaly Detection with Non-Local Information. in Advanced Analytics and Learning on Temporal Data, (2021), Springer, 186-194.
[4] Alexander Bronstein, et al. Partial similarity of objects, or how to compare a centaur to a horse, Int. J. Comput. Vis. 84(2), 163–183 (2009).
[5] Case School of Engineering: Bearing Data Center, Normal Baseline Data. Retrieved December 17, 2021 engineering.case.edu/bearingdatacenter/normal-baseline-data
[6] Hoang Anh Dau, et.al. (2019). The UCR time series archive. IEEE/CAA Journal of Automatica Sinica, 6(6), 1293–1305.
[7] Angus Dempster, et al. (2020). ROCKET: exceptionally fast and accurate time series classification using random convolutional kernels. Data Min. Knowl. Discov. 34(5): 1454-1495
[8] Hassan Ismail Fawaz, et al. "Inceptiontime: Finding alexnet for time series classification." Data Mining and Knowledge Discovery 34.6 (2020): 1936-1962.
[9] Germán H. Flores, Roberto Manduchi. "Weallwalk: An annotated dataset of inertial sensor time series from blind walkers." *ACM Transactions on Accessible Computing (TACCESS)* 11.1 (2018): 1-28.
[10] Ben D. Fulcher, et al. "Highly comparative time-series analysis: The empirical structure of time series and their methods." Journal of the Royal Society Interface 10.83 (2013): 20130048.
[11] Bryan Hooi, Shenghua Liu, Asim Smailagic, Christos Faloutsos: BeatLex: Summarizing and Forecasting Time Series with Patterns. ECML/PKDD (2) 2017: 3-19
[12] Kyle Hundman et al. Detecting Spacecraft Anomalies Using LSTMs and Nonparametric Dynamic Thresholding. ACM SIGKDD (2018), 387-395.
[13] Shima Imani, et al.: Matrix Profile XIII: Time Series Snippets: A New Primitive for Time Series Data Mining. ICBK 2018: 382-389
[14] IRENA (2019), Innovation landscape brief: Time-of-use tariffs, International Renewable Energy Agency, Abu Dhabi.
[15] Jason Lines, et al. "Time series classification with HIVE-COTE: The hierarchical vote collective of transformation-based ensembles." *ACM Transactions on Knowledge Discovery from Data* 12.5 (2018).
[16] Carl H. Lubba, et al. catch22: CAnonical Time-series CHaracteristics. Data Min Knowl Disc 33, 1821–1852 (2019).
[17] Laurens Van der Maaten and Geoffrey Hinton. "Visualizing data using t-SNE." JMLR 9, no. 11 (2008).
[18] Abdullah Mueen, et al. (2015) "The fastest similarity search algorithm for time series under Euclidean distance." url: www.cs.unm.edu/~mueen/FastestSimilaritySearch.html.
[19] Open Power System Data. 2020. Data Package Time series. Version doi.org/10.25832/time_series/2020-10-06.
[20] John Paparrizos, Luis Gravano. "Fast and Accurate Time-Series Clustering." ACM TODS 42, no. 2 (2017): 8.
[21] https://sites.google.com/view/precis-supplemental-materials
[22] Patrick Schäfer. "Scalable time series classification." Data Mining and Knowledge Discovery 30.5 (2016): 1273-1298.
[23] Ahmed Shifaz, et al. "TS-CHIEF: a scalable and accurate forest algorithm for time series classification." Data Mining and Knowledge Discovery 34.3 (2020): 742-775.
[24] Zhiguang Wang, et al. "Time series classification from scratch with deep neural networks: A strong baseline." *2017 International joint conference on neural networks (IJCNN)*. IEEE, 2017.
[25] Why Everything In Austria Is Closed On Sundays. Webpage retrieved June 2 2022. t24hs.com/why-everything-in-austria-is-closed-on-sundays-and-what-to-do-instead/
[26] Yeh, Chin-Chia Michael, et al. "Error-bounded Approximate Time Series Joins using Compact Dictionary Representations of Time Series." Proceedings of the 2022 SIAM International Conference on Data Mining (SDM). Society for Industrial and Applied Mathematics, 2022.
[27] Chin-Chia Michael Yeh, et al: Online Amnestic DTW to allow Real-Time Golden Batch Monitoring. KDD 2019: 2604-2612
[28] Chin-Chia Michael Yeh, et al. "Merchant Category Identification Using CreditCard Transactions." In 2020 IEEE International Conference on Big Data, pp. 1736-1744. IEEE, 2020.
[29] Mi Zhang and Alexander A. Sawchuk, "USC-HAD: A Daily Activity Dataset for Ubiquitous Activity Recognition Using Wearable Sensors", ACM International Conference on Ubiquitous Computing (UbiComp) (SAGAware), Pennsylvania, USA, 2012.